\documentclass{article}
\usepackage{spconf,amsmath,graphicx}
\usepackage{amsthm,amsmath,amssymb}
\usepackage{mathrsfs}
\usepackage[lined,boxed,commentsnumbered,ruled]{algorithm2e}
\usepackage{booktabs,multirow} 
\usepackage{graphicx}

\title{PSEUDO LABELS REFINEMENT WITH INTRA-CAMERA SIMILARITY FOR UNSUPERVISED PERSON RE-IDENTIFICATION}
\name{Pengna Li, Kangyi Wu, Sanping Zhou, Qianxin Huang, Jinjun Wang}
\address{Xi’an Jiaotong University\\Institute of Artificial Intelligence and Robotics\\28 West Xianning Road, Xi’an, Shaanxi, P. R. China}
\begin{document}
%\ninept
%
\maketitle
\begin{abstract}
Unsupervised person re-identification (Re-ID) aims to retrieve person images across cameras without any identity labels. Most clustering-based methods roughly divide image features into clusters and neglect the feature distribution noise caused by domain shifts among different cameras, leading to inevitable performance degradation. To address this challenge, we propose a novel label refinement framework with clustering intra-camera similarity. Intra-camera feature distribution pays more attention to the appearance of pedestrians and labels are more reliable. We conduct intra-camera training to get local clusters in each camera, respectively, and refine inter-camera clusters with local results. We hence train the Re-ID model with refined reliable pseudo labels in a self-paced way. Extensive experiments demonstrate that the proposed method surpasses state-of-the-art performance. 
\end{abstract}
\section{Introduction}
\label{sec:intro}

Person re-identification (Re-ID) is the task of retrieving images of a specified person under non-overlapping cameras \cite{zheng2016person}. Thanks to the development of deep learning technology, supervised Re-ID methods have led to great performance advances. Unfortunately, purely supervised methods expect a large amount of expensive labelled data, limiting the adaptability in the real world. Therefore, unsupervised settings have gained increasing research attention.

Unsupervised Re-ID works can be categorized into purely unsupervised learning methods and unsupervised domain adaption (UDA) methods based on whether using external labelled data. The former methods train the feature encoder without any identity labels while UDA methods require a labelled source domain and an unlabeled target domain. This paper addresses the purely unsupervised person Re-ID task, which is more scalable and challenging. Existing unsupervised works usually adopt pseudo-label-based methods, which alternate between generating pseudo labels by clustering \cite{lin2019bottom,dai2022cluster}, softened labels \cite{yu2019unsupervised,lin2020unsupervised} or label refinement \cite{cho2022part} and training the model with the generated labels. The inherent noises in pseudo labels severely degrade the performance of these unsupervised works, and recent studies address the challenge. MMCL \cite{wang2020unsupervised} trains the Re-ID model to recognize each image as multiple labels. SPCL \cite{ge2020self} proposes a unified contrastive learning framework to seek reliable labels. RLCC \cite{zhang2021refining} exploits cluster consensus to estimate pseudo-label similarity and refines pseudo-labels. Although previous methods have achieved state-of-the-art performance, most of them measured the similarity of image pairs without considering domain shifts among different cameras. Due to camera views and resolution gap, feature distribution of the same ID suffers from significant variance. Positive pairs captured from different cameras may have larger dissimilarity than negative samples from the same camera. Hence, it's difficult to identify image pairs of the same ID cross cameras and learn reliable intra-class similarity.

\begin{figure*}[t]
\centering
\includegraphics[width=14.5cm]{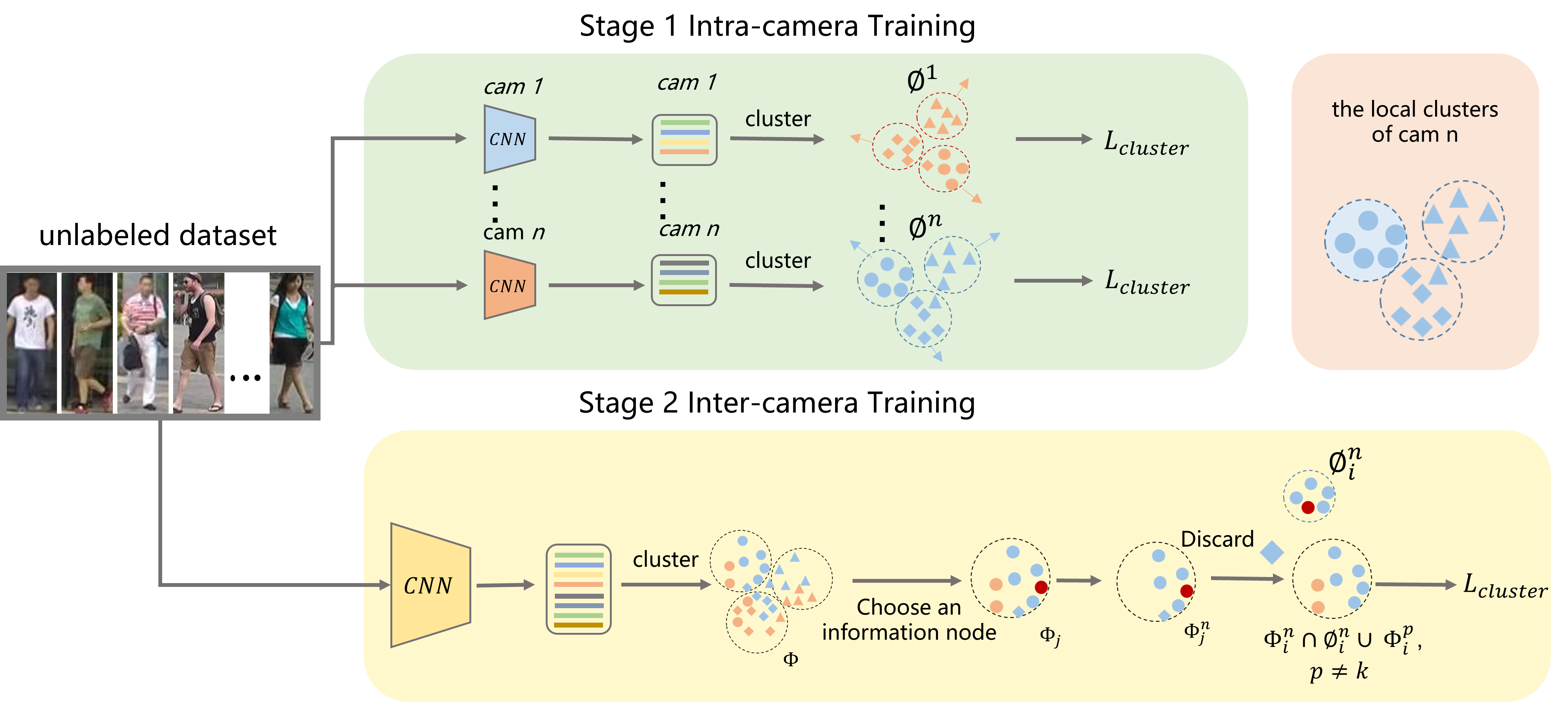}
\caption{The Illustration of our proposed method for unsupervised person Re-ID. The Intra-camera training stage extracts features and clusters to local clusters within each camera respectively. The Inter-camera training exploits the reliable local clusters results to refine inter-camera labels. The refined labels are adopted to optimize the inter-camera CNN.}
\label{1}
\end{figure*}

Recent works have utilized camera labels to  address the problem above. SSL \cite{lin2020unsupervised} proposes a cross-camera encouragement term to reduce intra-camera negative pairs by increasing the dissimilarity to image pairs under the same camera. CAP \cite{wang2021camera} splits each cluster into multiple camera-aware proxies according to the camera ID. The proxies can capture the local structure within the clusters and deal with the intra-ID differences and the inter-ID similarities. IICS \cite{xuan2021intra} alternates between intra-camera and inter-camera iterative training to optimize feature encoder. Compared to existing work that directly utilizes camera labels to optimize the re-ID model, this work is motivated to cluster intra-camera similarity to refine global inter-camera pseudo labels, and the refined labels are adopted to train the model.

In order to take full advantage of the inter-camera similarity, we propose a novel pseudo-label refinement framework to reduce label noise and exploit more reliable labels. Feature distribution in the same camera pays more attention to the appearance of pedestrians and could result in highly reliable intra-camera pseudo labels, which inspires us to make full use of intra-camera pseudo labels to refine noisy inter-camera pseudo labels. Specifically, we adopt a two-stage training strategy to optimize the model. The first stage is intra-camera training, which is conducted in each camera respectively. Features of each camera domain are separately extracted and clustered to generate independent local pseudo labels. The generated pseudo-labels are used as supervision to optimize their specific encoder. This stage ensures that the local pseudo labels are reliable enough to refine the global inter-camera clusters results in the next stage.

The second stage is inter-camera training conducted across cameras. Because of domain shifts among different cameras, global inter-camera label accuracy suffers greatly. Therefore, we utilize reliable intra-camera labels to handle the label noise. Specifically, we first choose some information nodes for each cluster, which usually have high utility. For each information node, query the relationship with the others in the same cluster and discard the negative samples based on cluster results of the first stage. The remaining pseudo-labels are more reliable for learning. In this way, we train the Re-ID model through self-paced learning. 

Our contributions can be summarized as follows. First, we exploit intra-camera similarity to effectively handle the label noise caused by the domain gap. Second, we propose a novel label refinement method to seek more reliable labels for feature learning. Besides, Extensive experimental results have demonstrated the state-of-the-art performance of our methods on multiple large-scale datasets.

\section{Methodology}
\label{sec:format}

\subsection{Problem Formulation}
\label{ssec:subhead}

We denote the unlabeled dataset as $\mathcal{D}=\{x_i\}_{i=1}^N$, where $x_i$ is a pedestrian image and $N$ is the total number of images. In our setting, the dataset could be divided into several domains based on camera information, $\mathcal{D}=\{\mathcal{D}^c\}_{c=1}^C$, where $c$ is camera ID, $C$ is the number of cameras. Besides, we denote $\mathcal{D}^c=\{x_i^c\}_{i=1}^{N^c}$, where $N^c$ is the number of images in camera $c$. Our goal is to learn a discriminative Re-ID model $f^\theta$ on $\mathcal{D}$. Given a query image $q$, the model $f^\theta$ is employed to extract discriminative features to retrieve images containing the person in the gallery $\mathcal{G}$. The Re-ID model is expected to predict similar features for images of the same person. 
To achieve this goal, we propose a novel unsupervised framework to refine labels with clustering intra-camera similarity. As illustrated in Fig.\ref{1}, we optimize $f^\theta$ by two stages of training, consisting of the intra-camera training stage which trains encoder $f^{\theta_c}$ and clusters the features respectively within each camera, and the inter-camera training stage which refines pseudo labels with intra-camera clusters and trains model $f^\theta$. The overall flowchart is illustrated in Algo.\ref{alg1}, and the details will be discussed in the rest of the section.

\subsection{Intra-camera training}
\label{ssec:subhead}

Given the unlabeled images with camera information, the intra-camera step aims to obtain reliable local intra-camera pseudo-labels. Specifically, we adopt a pre-trained model to extract features for input images and perform clustering in each camera domain $\mathcal{D}^c$ separately. Images within the same cluster are assigned identical labels. With the generated pseudo labels, the intra-camera training stage could learn a specific feature encoder $f^{\theta_c}$ for each camera. The feature distribution in each independent camera is not affected by the camera domain gap and focuses on the characteristics of the person's identity. Hence, the model training can have less performance degradation from label noise. The intra-camera step can reduce intra-ID noise and provide more reliable local pseudo labels. Stage 1 saves the final clustering results $\phi^c$ for the latter label refinement.

\subsection{Inter-camera training}
\label{ssec:subhead}

Inter-camera step follows the intra-camera pipeline to cluster the global features and assign images with identity labels across cameras. The feature encoder $f^\theta$ pre-trained on ImageNet \cite{deng2009imagenet} can learn general feature representations. But it's too hard to gain reliable pseudo-labels in the early epochs. The model is expected to learn from simple and reliable samples and then gradually learn complex and hard samples in a self-paced way. Therefore, we exploit intra-camera clustering results as prior knowledge to refine the global pseudo labels. Due to the limited nature of intra-camera clustering, we can only query image pair relationships under the same camera. At the beginning of the training procedure, each global cluster may contain a majority of nodes having the same camera ID, in which some impurities are incorrectly clustered. Our label refinement aims to discard these impurities that tend to degrade the performance of the model. If we can find information nodes having high utility in each cluster, the rest is naive. We only require to ask for the local clustering results to compare the information node with other clustered nodes and discard the negative sample. Intuitively, information nodes are expected to be the centre of the cluster and be close to others clustered nodes. We design a criterion to estimate the centrality based on Harmonic Centrality \cite{marchiori2000harmony} as follows: 
\begin{equation}
    score(i)=\sum_{j\in{{top}_{15}(i)}}\frac{1}{dist(i,j)+mean(dist)}
    \label{e2}
\end{equation}
where $dist(,)$ denotes the distance between $i$ and $j$, and $mean(dist)$ is the mean of all distance. ${top}_{15}(i)$ is top 15 closest neighbors of $i$. The higher $score(i)$ is, the more utility $i$ has. Hence, we choose samples with $score$ higher than the mean of all $score$ as the information nodes, which could dynamically update in each epoch.

Given an information node $i$ in camera $c$, find the local cluster $\phi^c_i$ and the global cluster $\Phi_i$ where the information node is located. The global cluster $\Phi_i$ can be divided into multiple sub-clusters $\{\Phi_i^c\}$ in a per-camera manner. For each node in $\Phi_i^c$, we query the relationship between the node and $i$ according to intra-camera clustering results. We keep the positive samples and discard the negative samples with a certain probability $p$. When we set the probability $p=1$, the refined global cluster is defined as: 
\begin{equation}
    \Phi_i={\Phi_i^c}\cap{\phi_i^c}\cup{\Phi_i^p}
    \label{e3}
\end{equation} where $p$ is the other camera IDs, $p\neq c$. We exploit the refine clusters to train feature encoder $f^\theta$. As mentioned above, the model is expected to train from the first reliable and simple samples to complex and hard samples. as the training continues, we could make the samples with refined labels more complex by decaying the probability. In this way, we can train the model with self-paced learning.

\subsection{Loss function}
\label{ssec:subhead}
In our framework, we employ ClusterNCE loss \cite{dai2022cluster} for model optimization. For intra-inter-camera training, the loss can be defined as follows: 
\begin{equation}
    \mathcal{L}_{cluster}=-log\frac{exp(q\cdot{u_+}/\tau)}{\sum_{k=0}^{K}exp(q\cdot{u_k}/\tau)}
    \label{e1}
\end{equation}
where $q$ is the query feature and $u_k$ is the cluster centroid defined by the mean feature vectors of each cluster. $u_+$ shares the same pseudo label with the query. $\tau$ is a temperature hyper-parameter. All cluster features representation can be stored in a memory dictionary, which is updated consistently by corresponding query $q$ as:
\begin{equation}
    u_k = mu_k+(1-m)q
    \label{e4}
\end{equation}
where $m$ is the momentum updating factor.  The detailed training setting is described in Sec.\ref{ssec:setting}

\begin{algorithm}[t]
\SetAlgoLined
\KwIn{An unlabeled dataset $\mathcal{D}$, a CNN model $f^\theta$ pre-trained with ImageNet}%输入参数
\KwOut{Optimized model $f^\theta$}%输出
% \KwResult{Write here the result}
Stage 1: intra-camera training
\\
\For{$c=1,...,N^c$}{
Initialize $f^{\theta_c}$ with $f^\theta$
\\
\For{$epoch = 1,...,nums\_epochs$}{
Extract features for $\mathcal{D}^c$ with $f^{\theta_c}$
\\
Clustering features into local clusters set $\phi^c$
\\
Training with local pseudo labels
}
Save the final clustering results $\phi^c$
}
Stage 2: Inter-camera training
\\
\For{$epoch = 1,...,nums\_epochs$}{
Extract features for $\mathcal{D}$ with $f^{\theta}$
\\
Clustering features into global clusters $\Phi$
\\
Choose information nodes with Eq.\ref{e2}
\\
\For{i in information node}{
Find global cluster $\Phi_{i}$, camera ID $c$ and local cluster $\phi_{i}^c$ of node, denote $p\neq c$
\\
Refine pseudo labels with Eq.\ref{e3}\\
}
Training with the refined global pseudo labels
}

\caption{Our pipeline with label refinement}
\label{alg1}
\end{algorithm}

\section{Experiments}
\label{sec:format}

\subsection{Datasets and Settings}
\label{ssec:setting}
We evaluate our proposed method on Market1501 \cite{zheng2015scalable} and MSMT17 \cite{wei2018person} respectively. Market1501 contains 32668 pedestrian images of 1501 pedestrian identities captured by 6 non-overlapping cameras. MSMT17 adopts 15 cameras to collect data, including 126441 images of 4101 person IDs. We adopt two commonly used evaluation metrics, \text{i.e,}, Cumulative Matching Characteristic (CMC) at Rank-k, and mean average precision (mAP) to evaluate our method. 

The implementation of the framework adopts 4 GTX-2080TI GPUs for training. The model modification follows Cluster-Contrast \cite{dai2022cluster}. We train the model for 20 epochs at the intra-camera step and 50 epochs at the inter-camera step. At the beginning of the epoch, we perform the Agglomerative Hierarchical \cite{pedregosa2011scikit} clustering. For Market1501, the number of clusters is 600 at intra-camera training and 800 at inter-camera training. For MSMT17, the number of clusters is $num{\_}images/5$ for each camera at the intra-camera stage and 1200 at the inter-camera stage.
\begin{table}[h]
\vspace{-1.0em}
	\caption{\textbf{Comparison with ReID methods}}
        \label{table1}
	\centering                                       %把表居中
    \resizebox{\linewidth}{!}	{
    \begin{tabular}{cccccc}
		\toprule                                     %第一道横线
		\multirow{2}{*}{Method}&
            \multirow{2}{*}{Venue}&
		\multicolumn{2}{c}{Market1501}& 
		\multicolumn{2}{c}{MSMT17}\cr     % \cr表示回车
		\cmidrule{3-4}\cmidrule{5-6}
		&&mAP &Rank-1 &mAP &Rank-1 \\
		\midrule                                      %第二道横线
            \multicolumn{6}{l}{Unsupervised methods without camera labels}\\
            \midrule
		\multicolumn{1}{c}{BUC\cite{lin2019bottom}}& AAAI'19& 38.3& 66.2& -& - \\
            \multicolumn{1}{c}{MMCL\cite{wang2020unsupervised}}& CVPR'20& 45.5& 80.3& 11.2& 49.8 \\
            \multicolumn{1}{c}{SPCL\cite{ge2020self}}& NeurIPS'20& 73.1& 88.1& 19.1& 42.3 \\
            \multicolumn{1}{c}{RLCC\cite{zhang2021refining}}& CVPR'21& 77.7& 90.8& 27.9& 56.5 \\
            \multicolumn{1}{c}{ICE\cite{chen2021ice}}& ICCV'21& 79.5& 92.0& 30.3& 60.8 \\
            \midrule                                      %第二道横线
            \multicolumn{6}{l}{Unsupervised methods with camera labels}\\
            \midrule
		\multicolumn{1}{c}{SSL\cite{lin2020unsupervised}}& CVPR'20& 37.8& 71.7& -& - \\
            \multicolumn{1}{c}{CAP\cite{wang2021camera}}& AAAI'21& 79.2& 91.4& 36.9& 67.4 \\
            \multicolumn{1}{c}{IICS\cite{xuan2021intra}}& CVPR'21& 72.9& 89.5& 26.9& 56.4 \\
            \multicolumn{1}{c}{ICE\cite{chen2021ice}}& ICCV'21& 82.3& \pmb{93.8}& 38.9& 70.2 \\
            Ours& {This work}& \pmb{83.2}& 93.1& \pmb{43.3}& \pmb{71.5}\\
		\bottomrule                                   %第三道横线
	\end{tabular}
    }
    \vspace{-1.0em}
\end{table}

\begin{table}[h]
\vspace{-1.0em}
	\caption{\textbf{Ablation study on decay methods}}
        \label{table2}
	\centering                                       %把表居中
    \resizebox{\linewidth}{!}{
    \begin{tabular}{cccccc}
        \toprule                                     %第一道横线
		\multirow{2}{*}{Method}&
        \multirow{2}{*}{Decay}&
		\multicolumn{2}{c}{Market1501}& 
		\multicolumn{2}{c}{MSMT17}\cr     % \cr表示回车
		\cmidrule{3-4}\cmidrule{5-6}
		&&mAP &Rank-1 &mAP &Rank-1\\
		\midrule                                      %第二道横线
        Baseline& -& 81.8& 92.9& 33.0& 61.4 \\
        \midrule 
        Ours& -& 82.5& 92.7& 39.8& 68.5\\
        Ours& Linear& 81.3& 91.9& 41.3& 69.7\\
        Ours& Exponential& 83.1& 92.6& 41.4& 69.3\\
        Ours& Cosine& \pmb{83.2}& \pmb{93.1}& \pmb{43.3}& \pmb{71.5}\\
		\bottomrule  %第三道横线
	\end{tabular}
    }
    \vspace{-1.0em}
\end{table}

\subsection{Comparison with State-of-the-Arts}
\label{ssec:subheading}

We compare our proposed method with state-of-the-art unsupervised Re-ID methods on Market-1501 and MSMT17. All the results are in Table \ref{table1}. These Re-ID methods are categorized into unsupervised methods that do not use camera labels and unsupervised methods trained with camera labels. Our method obtains 83.2 \% mAP and 93.1\% Rank-1 on Market1501 and 43.3 \% mAP and 71.5 \% Rank-1 on MSMT17, which proves the effectiveness of our method for large-scale complex datasets. CAP \cite{wang2021camera} and IICS \cite{xuan2021intra} utilize intra-camera similarity to generate pseudo labels, which are directly used for optimizing the Re-ID model to eliminate intra-ID variance. Differently from these methods, we exploit the reliable intra-camera pseudo labels to refine inter-camera labels and train the model in a self-paced way. Experiments distinctively demonstrate the superior performance of our proposed method.

\begin{figure}[t]
\centering
\includegraphics[height = 2.8cm, width = \linewidth]{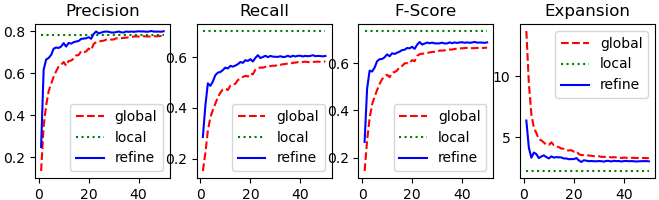}
\caption{The Illustration of clustering quality on MSMT17.}
\label{img1}
\vspace{-1.0em}
\end{figure}

\subsection{Ablation study}
\label{ssec:subheading}

In order to verify the necessity of decaying the probability of discarding negative samples, we conduct ablation experiments. All results are summarized in Table \ref{table2}. Compared with the baseline model, our label refinement method significantly boosts performance. The decay strategy is proposed to gradually improve the training samples' complexity. The models using decay strategies achieve superior performance, which validates the effectiveness of the strategy. Our model performs best when adopting the cosine decay strategy.

We also evaluate the clustering quality  of global, local, and refined clusters over training epochs on MSMT17 in Fig.\ref{img1}. We employ precision, recall, f-score, and expansion to analyze the clustering quality. The model doesn't use any decay strategies. Fig.\ref{img1} demonstrates that the quality of refined clusters is superior to the quality of global clusters, which verifies the validity of our pseudo labels refinement.

\section{Conclusion}
\label{sec:format}

In this work, we propose a novel label refinement framework with clustering intra-camera similarity for unsupervised person Re-ID. Our method aims to handle the intra-identity variance caused by different camera views. We conduct intra-camera training to obtain reliable local clusters. In order to effectively refine labels, we design a criterion to select information nodes to refine global clusters with local results at the inter-camera stage. Extensive experiments show the effectiveness and efficiency of our method.

\vfill\pagebreak

\bibliographystyle{IEEEbib}
\bibliography{strings}
\end{document}